\pdfoutput=1
\documentclass[11pt]{article}

\usepackage[]{EMNLP2022}

\usepackage{times}
\usepackage{latexsym}
\usepackage{booktabs}

\usepackage{amsmath}
\usepackage{amsthm}
\usepackage{amssymb}
\usepackage{amsfonts}
\usepackage{mathabx}

\usepackage[T1]{fontenc}
\usepackage[utf8]{inputenc}

\usepackage{microtype}

\usepackage{inconsolata}

\newcommand{\mysubsection}[1]{\vspace{0.3em}\noindent\textbf{#1}}
\usepackage{graphicx}
\usepackage{pgfplots}
\usepackage{filecontents}

\title{Unifying Data Perspectivism and Personalization:\\An Application to Social Norms}

\author{Joan Plepi $\dagger$ \and Béla Neuendorf $\dagger$ \and Lucie Flek $\dagger \ddagger$ \and Charles Welch $\dagger \ddagger$\\
     Conversational AI and Social Analytics (CAISA) Lab \\ 
    $\dagger$ Department of Mathematics and Computer Science, University of Marburg \\ 
    $\ddagger$ The Hessian Center for Artificial Intelligence (Hessian.AI) \\
    \texttt{\{plepi,neuendob,lucie.flek,welchc\}@uni-marburg.de} 
} 

\begin{document}
\maketitle
\begin{abstract}
Instead of using a single ground truth for language processing tasks, several recent studies have examined how to represent and predict the labels of the set of annotators. However, often little or no information about annotators is known, or the set of annotators is small. In this work, we examine a corpus of social media posts about conflict from a set of 13k annotators and 210k judgements of social norms. We provide a novel experimental setup that applies personalization methods to the modeling of annotators and compare their effectiveness for predicting the perception of social norms. We further provide an analysis of performance across subsets of social situations that vary by the closeness of the relationship between parties in conflict, and assess where personalization helps the most.

% Methods in the natural language processing community have been developed to improve model performance by accounting for variables that pertain to subpopulations or individual users. We investigate the effect of these personalized representations for the task of predicting a users perception of social norms and discuss the how the impact of personalization is moderated by properties of the classification task.
\end{abstract}

% Two items on terminology to think about
% 1. do we refer to people as users or authors? we should choose one and I kind of like author better because a user can do many things but 'author' implies that they wrote something
% 2. do we refer to the clusters as "tasks"?
\section{Introduction}
Obtaining a single ground truth is not possible or necessary for subjective natural language classification tasks~\cite{ovesdotter-alm-2011-subjective}. Each annotator is a person with their own feelings, thoughts, experiences, and perspectives~\cite{basile2021we}. In fact, researchers have been calling for the release of data without an aggregated ground truth, and for evaluation that takes individual perspectives into account~\cite{flek-2020-returning}.

%LF: cite also the Lrec workshop here maybe?

% \cite{prabhakaran-etal-2021-releasing} -- aggregation unfairly removes the perspectives of sociodemographic groups
% \cite{aroyo2015truth} -- multiple truths and disagreement is good
% \cite{basile2020s}\cite{basile2021we} -- these two call for preaggregated data in released data and in evaluation
The idea that each annotator has their own view of subjective tasks, and even those previously thought to be objective was introduced by \citet{basile2020s} as \textit{data perspectivism}. A growth in the interest of this viewpoint has led to the 1st Workshop on Perspectivist Approaches to NLP in 2022.
Work has examined how to model annotators for subjective tasks and to predict each annotator's label~\cite{davani2021dealing,fornaciari-etal-2021-beyond}.
%LF: Not previous if it comes after? maybe cite earlier stuff like ruder plank?
Modeling annotator perspectives requires the release of corpora that include annotator-level labels rather than aggregated ``ground truth'' labels. \citet{bender2018data} further recommend releasing data statements that describe characteristics including who is represented in the data and the demographics of annotators. Such information is beneficial for raising awareness of the biases in our data. %cite Bender dataset guidelines?
While some corpora contain this information, like those for humor, emotion recognition, and hateful or offensive language, they contain few annotators and no additional information about them~\cite{meaney-etal-2021-semeval,kennedy2018gab,demszky-etal-2020-goemotions}.

An additional complication for subjective tasks is the fact that different people will interpret text in different ways. What is deemed toxic or offensive depends on who you ask~\cite{sap2021annotators,leonardelli-etal-2021-agreeing}. There are notable differences in perceived and intended sarcasm~\cite{oprea-magdy-2019-exploring,plepi-flek-2021-perceived-intended-sarcasm}. How one perceives the receptiveness of their own text is different than how others see it~\cite{yeomans2020conversational}. For such tasks, predicting the label given by third party annotators,  without knowing much about them, is not very useful. Modeling annotators with  personalization methods requires a corpus with many self-reported labels from many annotators, and additional contextual information about them.
% In addition to annotator label sets, we have the text that users wrote when giving their labels as well as other posts they have made on Reddit.

\begin{figure}
    \centering
    \includegraphics[width=0.45\textwidth]{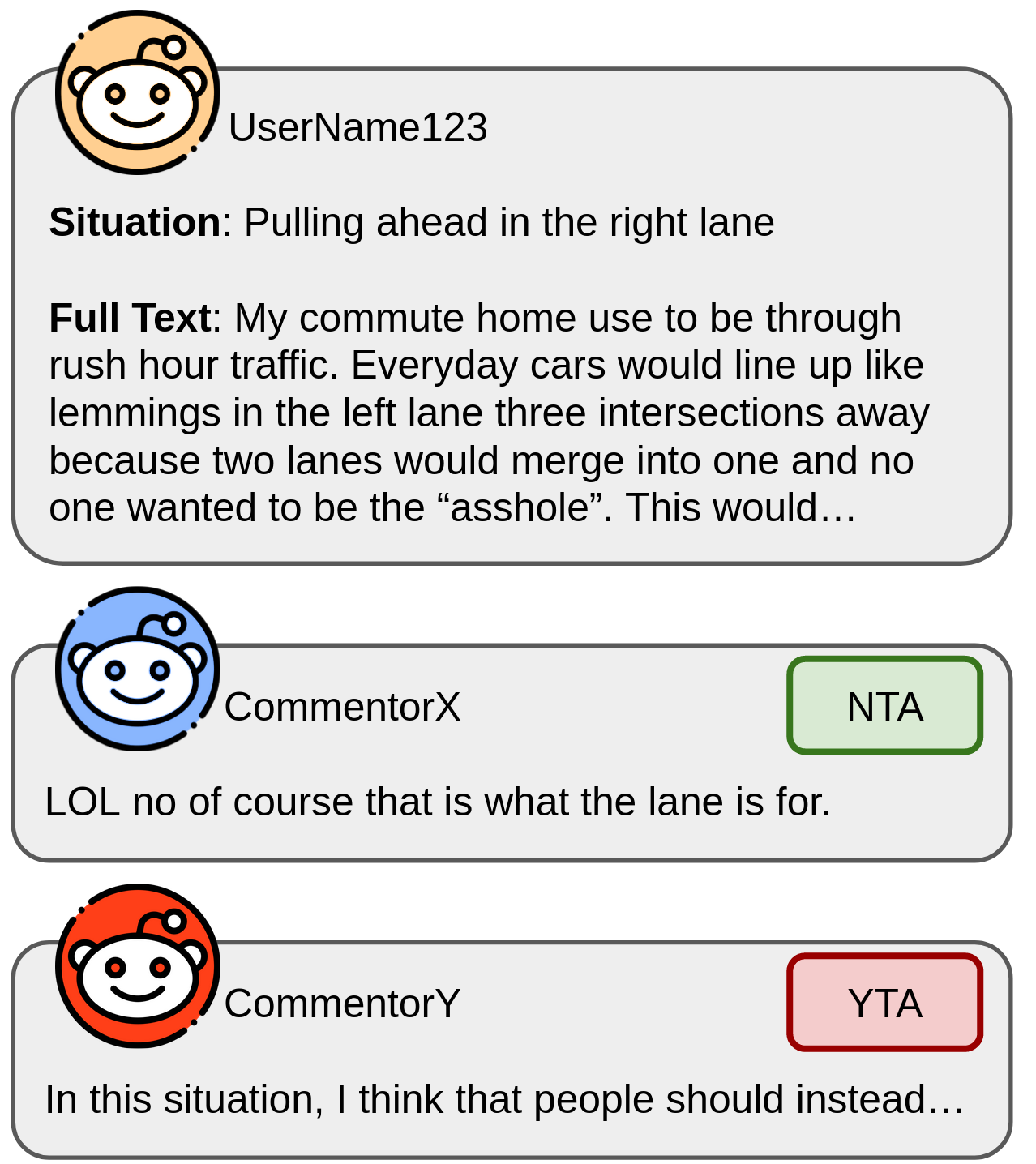}
    \caption{Example of a post on Reddit and two comments. The post has the situation, which comes from the post title and the full text of the post (truncated here). Usernames appear next to the icons of the poster and commentors. Each comment has a verdict, which is the label they assign (YTA or NTA).}
    \label{fig:reddit_example}
\end{figure}

In this work, we use English textual data in the form of posts from the website, Reddit, about social norms from the subreddit \texttt{/r/amitheasshole} (AITA). As shown in Figure~\ref{fig:reddit_example}, users of this online community post descriptions of situations, often involving interpersonal conflict, and ask other users to judge whether the user acted wrongly in the situation or not. %LF: this would be the first instance of the user vs author termino gy issue...
The judgements from these users constitute our labels, and their authors are the set of annotators (and we refer to them as such for the remainder of the paper), which allows us to explore methods to model annotators at a larger scale.
We explore methods of personalization to model these annotators and examine how the effectiveness of our approach varies with the social relation between the poster and others in the described situation. We further provide an analysis of how personalization affects demographic groups and how performance varies across individuals.

Our contributions include (1) a discussion of the relation between data perspectivism and personalization, (2) a novel problem setting involving a recently collected dataset with unique properties allowing us to explore these concepts for annotator modeling, and (3) a novel comparison of contemporary personalization methods in this setting.

% TODO more about novelty -- first approach to the best of our knowledge to model annotators on a large scale using additional information from the annotators other than their labels and per-instance text. First comparison of contemporary personalization methods in this setting.

\section{Formulation}\label{sec:formulation}
We formalize our task in terms of the textual data points, their authors, annotators, and the annotations they provide. A poster, $u$, makes a post, $p$, which is then commented on by an annotator, with ID $a$, who provides a comment, $c_{a,p}$, and a label, or verdict, $v_{a,p}$. Since we are modeling annotators, $u$ is not important to us, except that $u\ne a$ within the same post. Each post $p$ has many comments $c_p$, though this is not strictly necessary for our purposes, it does help reveal the subjectivity of the task. Importantly, each annotator, $a$, has many comments, $c_a$. In our case, the comment $c_{a,p_i}$ written by annotator $a$ on the $i$-th post $p_{i}$, is linked to a single $v_{a,p_i}$, though one could gather these from separate sources, and doing so may be necessary for other corpora. The subjective nature of the task and its evaluation comes from the assumption that annotators provide different verdicts for a post. %, $v_{p,i}$

Work on annotator modeling attempts to estimate the probability of a verdict given the post and annotator, $p(v_{a,p}|a,p)$. This is in contrast to predicting what an individual's language means,  $p(v_{a,p}|a,p,c_{a,p})$, which we refer to as a personalized classification task. Importantly, we make this distinction because personalization has historically focused on predicting a label assigned to an individual's text in a particular context (e.g. the sentiment of a review), whereas work on modeling annotators focuses on the label an individual \textit{would assign} in that context.

There is often no information about annotators or only an ID is known.
A few works on annotator modeling include extra information about the annotator, $T$. This information can be defined in various ways (see \S\ref{sec:annotator_disagree}). In this work, we use a collection of other texts from the annotator (see \S\ref{sec:methods}). To the best of our knowledge, our formulation of $T$ is novel in that it allows the application of previously developed methods for personalization to the task of annotator modeling.
% but for our task we use other comments from the annotator, $T=\{c_{a,q}|q\ne p, \forall q \in P\}$, where $P$ is the set of posts the annotator has commented on.
Importantly, we are predicting how the annotator will label the post, $p(v_{a,p}|a,p,T)$, not how to interpret their text. For other work that has attempted to interpret verdicts, $p(v_{a,p}|a,c_{a,p})$, refer to \S\ref{sec:social_norms}.
% The T and the p are optional for interpreting verdicts / personalization but I'm kind of surprised to see that most work only uses a and c.

% \vskip -0.3in

\section{Related Work}\label{sec:related_work}
In our work, we refer to users as social media users. Within AITA, we refer to the poster as the user who originally made the post, and annotators as those who commented on the post. Both posters and annotators are authors of their respective posts and comments.

%LF: this chapter is a shortening candidate to distill only the relevant stuff for the paper's contributions
\subsection{Social Norms}\label{sec:social_norms}
\citet{lourie2020scruples} looked at the AITA subreddit to model judgements of social norms. They looked at how to predict the distribution of judgements for a given situation, which indicates how controversial a situation may be. \citet{forbes-etal-2020-social} expanded on this study by using their data to extract a corpus of rules-of-thumb for social norms. We examine a new dataset, created from the posts in their data but including the set of comments, which include annotators, their label, and the accompanied comment~\cite{welch2022understanding}.

\citet{efstathiadis-etal-2021-explainable} examined the classification of verdicts at both the post and comment levels, finding that posts were more difficult to classify. \citet{botzer-etal-2022-analysis} also constructed a classifier to predict the verdict given the text from a comment and used it to study the behavior of users in different subreddits.
% They looked at the age and gender of posters and found minimal effect on the judgements received by those users. In contrast, \citet{deCandia-2021-modeling} found that younger and male (independently) users receive more negative judgements.
\citet{deCandia-2021-modeling} found that the subreddits where a user has previously posted can help predict how they will assign judgements. The author manually categorized posts into five categories: family, friendships, work, society, and romantic relationships. They found that posts about society, defined as ``any situation concerning politics, racism or gender questions,'' were the most controversial.
Several works have also looked at the demographic factors or framing of posts affects received judgements~\cite{zhou-etal-2021-assessing,deCandia-2021-modeling,botzer-etal-2022-analysis}.
% \citet{zhou-etal-2021-assessing} found that the use of passive voice and framing as an \textit{agent} or \textit{patient} has an effect on judgement received.
%LF: In contrast, our work something something...?

\subsection{Personalization}
Many different approaches and tasks have used some form of personalization. These methods use demographic factors~\cite{hovy-2015-demographic}, personality traits~\cite{lynn-etal-2017-human}, extra-linguistic information that could include context, or community factors~\cite{bamman2015contextualized}, or previously written text. A similarity between personalization and annotator modeling is that the most common approach appears to be using author IDs. These have been used, for instance, in sentiment analysis~\cite{mireshghallah2021useridentifier}, sarcasm detection~\cite{kolchinski-potts-2018-representing}, and query auto-completion~\cite{jaech-ostendorf-2018-personalized}.

\citet{king-cook-2020-evaluating} evaluated methods of personalized language modeling, including priming, interpolation, and fine-tuning of n-gram and neural language models. \citet{wu-etal-2020-ptum} modeled users by predicting their behaviors online. Similarly, one's use of language can be viewed as a behavior. \citet{welch-etal-2020-exploring} modeled users by learning separate embedding matrices for each user in a shared embedding space. \citet{welch-etal-2022-leveraging} explored how to model users based on their similarity to others. They used the perplexity of personalized models and the predictions of an authorship attribution classifier to generate user representations. In social media in particular, a community graph structure can be used to model relationships between users and their linguistic patterns~\cite{yang2017overcoming}.

\subsection{Annotator Disagreement}\label{sec:annotator_disagree}
There has been a shift in thinking about annotator disagreement as positive rather than negative~\cite{aroyo2015truth}. Disagreement between annotators is often resolved through majority voting~\cite{nowak2010reliable}. In some cases, label averaging can be used~\cite{sabou-etal-2014-corpus}, or disagreements can be resolved through adjudication~\cite{waseem-hovy-2016-hateful}. Majority voting, which is most often used, takes away the voice of underrepresented groups in a set of annotators, for instance, older crowd workers~\cite{diaz2018addressing}, and aggregation in general obscures the causes of lower model performance and removes the perspectives of certain sociodemographic groups~\cite{prabhakaran-etal-2021-releasing}. On the other hand, \citet{geva-etal-2019-modeling} uses annotator's identifiers as features to improve model performance while training. They note that annotator bias is a factor which needs additional thought when creating a dataset.

\citet{fornaciari-etal-2021-beyond} predict soft-labels for each annotator to model disagreement which mitigates overfitting and improves performance on aggregated labels across tasks, including less subjective tasks like part-of-speech tagging. \citet{davani2021dealing} developed a multi-task model to predict all annotator's judgements, finding that this achieves similar or better performance to models trained on majority vote labels. They note that a model that predicts multiple labels can also be used to measure uncertainty. They experiment with two datasets, which have fewer than a hundred annotators each. This allows them to model all annotators, though they note that training their model on corpora with thousands of annotators, like ours, is not computationally viable.

Most work models annotators using their ID only. \citet{basile2021toward} has called for extra information about annotators to be taken into account. Some annotation tasks have collected demographic information about annotators, for instance~\cite{sap2021annotators}, or used the confidence of annotators as extra information~\cite{cabitza2020if}.

% It is also important to note that in our work, we are modeling the person who wrote the text we are trying to classify. In work on annotator disagreement, if the annotator is to be modeled, they are an outside observer interpreting someones text. To the best of our knowledge, our work is novel in exploring approaches to personalizing models of first party annotated subjective data.

\section{Dataset}
%TODO: Social norms + our verdicts extraction, how we get demographics, what are the splits, what is the distribution

We use the dataset of \cite{welch2022understanding}, who collected data from Reddit, an online platform with many separate, focused communities called subreddits. The data is from the AITA subreddit, where members describe a social situation they are involved in, and ask members of the community for their opinions. Others then decide if the poster is the wrongdoer in the situation, in which case they will respond with ``you're the asshole'' (YTA), or ``not the asshole'' (NTA). The dataset uses the posts from \citet{forbes-etal-2020-social}, and also includes the post title, full text, all comments, and their corresponding authors. %that contain a verdict of YTA or NTA.\footnote{Reddit posts are crawled with the Reddit API (\url{https://www.reddit.com/dev/api}) and comments with the PushShift API (\url{https://files.pushshift.io/reddit/comments/}).}
The comments are preprocessed in order to extract the ones that contain a verdict of YTA or NTA,\footnote{Reddit posts were crawled with the Reddit API (\url{https://www.reddit.com/dev/api}) and comments with the PushShift API (\url{https://files.pushshift.io/reddit/comments/}).} and others were removed. In order to extract verdicts, they manually created a set of keywords for both classes, and filtered the comments to remove these expressions.

They also crawled the historical posts of the annotators who have commented. The initial dataset contains 21K posts,  364K verdicts (254K NTA, 110K YTA) written by 104K different authors.
In our experiments, we keep only the annotators with more than 5 verdicts. 
This results in 210K verdicts (150K NTA, 60K YTA) written from 13K different annotators.

\subsection{Extracting Demographics}\label{sec:extract_demos}
We modify the script from \citet{welch-etal-2020-compositional}, which extracts age and gender using a set of phrases such as ``I am a woman'' and ``I am X years old'', to also capture Reddit shorthand. Reddit's users often disclose their age and gender when telling stories or asking for help. This often takes the form of a letter and number in brackets or parentheses (e.g. ``[32F]'' for age 32 and female), immediately after a first person pronoun. We base this extraction on recent work that has used similar methods~\cite{botzer-etal-2022-analysis,deCandia-2021-modeling}.
Additionally, we capture gender expressed in a phrase containing an adjective, such as ``I am a quiet man''. 
We adjusted the regex to exclude false positives like ``I am a manager'' or %emographic information included in quoted text and to label phrases like 
``I am a manly girl''. %correctly, which methods above labeled as male. Even after excluding those, we find that we are able to extract more ages and genders than with the methods above, i.e. 118\% more ages, 5\% more genders and 94\% more with both for the authors in our dataset, compared to \citet{welch-etal-2020-compositional}.

We then split ages into two groups. The median age of 28 is used to group people into \textit{younger} and \textit{older}. The resulting dataset contains 1,121 younger people (8\% of total) and 1,032 older (7.6\%). For gender, we find 2,280 are male (16.8\%) and 3,392 female (25\%). Note that our scripts exclude many people, including those who are non-binary. See our limitations section for more details.

\section{Clustering}\label{sec:clustering}

For the purpose of this work, we cluster the types of situations in our dataset as described at \cite{welch2022understanding}. To obtain this clustering, a graph with nodes representing the posts (situations) and edges weighted by their similarity is constructed. The situations are represented by their post titles or full text, and the relevant text is embedded with Sentence-BERT  (SBERT, \citet{reimers-gurevych-2019-sentence}) to form the set of nodes of the graph. The edges are created for each pair of situations using the cosine similarity normalized to $[0,1]$, resulting in a fully connected weighted graph.

Furthermore, the Louvain clustering algorithm is applied to obtain the clusters, which maximizes the modularity of the weighted graph~\cite{blondel2008fast}. Afterwards, an algorithm is applied to prune the graph. The N\% lowest weighted edges of the graph in steps of 10\% are dropped, and Louvain clustering is applied at each step. Then, for each pair of graphs that differ only by one step, that is having X\% and (X+10)\% of edges removed, the adjusted rand index is determined between those graphs to find a range where the clustering is least affected by edge removal and thus persistent. This method leads to a 30\% cutoff for full texts and 40\% for situations. In both cases, this resulted in 3 clusters which were then labeled by two annotators. 

\section{Methodology}\label{sec:methods}

% Given the situation, author and verdict, our goal is to predict the moral judgment of the verdict (NTA or YTA). We define the problem formally as follows. Given a verdict $v \in V$, with moral judgement $y \in \{+1, -1\}$ and a situation $s \in S$, we want to maximize the conditional probability $p(y | v, s)$. Moreover, given a user $u \in U$, we want to evaluate the influence in predicting the moral judgement, hence maximizing the conditional probability $p(y | v, s, a)$. In the following subsections we describe how we model different information, like text and users.
As described in \S\ref{sec:formulation}, we are attempting to model $p(v_{a,p}|a,p,T)$, or the probability of a verdict given by an annotator of a post, with additional information about that annotator, $T$. In our work, we define $T=\{c_{a,q}|q\ne p, \forall q \in P\}$, where $P$ is the set of posts the annotator has commented on. In the following subsections, we describe how we encode text and the personalization methods we use to model $T$.

\subsection{Encoding Text}
For the purpose of our experiments, we first extract initial representations for our textual information, which may be either a situation or a verdict. We utilize SBERT embeddings to encode texts. SBERT is specifically pretrained, such that it is able to produce semantically meaningful sentence representations. % We can add here some works where it has proven successfully
Formally, given a text $t$ we have,
    $\overline{t} = SBERT(t)$ 
where $ \overline{t} \in \mathbb{R}^{d} $, and $SBERT$ computes the sentence representation.
\subsection{Encoding Annotators}
In this section, we describe various personalization methods that we use to represent the annotators in our dataset.

\mysubsection{Sentence BERT for Annotators.}\label{subsec:avg} 
Given an annotator, $a$, with the set of comments $T$, where $ |T| = n_a $, we average the SBERT embeddings $\overline{c}_{a, q_{k}}$ of their historical posts to compute the final annotator representation as follows: $\overline{a} = \frac{1}{n_a} \sum_{k=1}^{n_a} \overline{c}_{a,q_{k}}$.

\mysubsection{Priming.} This method, originally used in recurrent neural networks, passes data from a given annotator through the model to alter the parameters before passing the text to use for prediction~\cite{king-cook-2020-evaluating,lee2020misinformation}.
% Another personalization method that we use is priming similar to the work on \cite{king-cook-2020-evaluating} \cite{lee2020misinformation}.
In our work, we also sample annotator data, but instead append it to the text to classify.
For every annotator $a$, we randomly sample a number of comments from their context $T$, until the maximum number of tokens is less than $m$. The sampled text for each annotator is concatenated to the beginning of the input text that is being classified during fine-tuning.\footnote{We also tried concatenating to the end in preliminary experiments, though we found performance was slightly lower.}

\mysubsection{Authorship Attribution.} In the authorship attribution method, we model a task $p(a | c_{a,p})$, that is, given a comment we want to predict the author. We use SBERT to extract the initial embeddings of the text, and we forward these representations into a two-layer feed-forward network parameterized from weight matrices $W_{1} \in  \mathbb{R}^{\frac{d}{2} \times d}$ and $W_{2} \in  \mathbb{R}^{n \times  \frac{d}{2}}$, where $d$ is the dimension of the SBERT embeddings, and $n$ is equal to the number of annotators during the training. The output of the last linear layer is then passed to a softmax layer to get a distribution over the authors. After training, we use the linear layers to extract the initial annotators' representations. For each annotator $a$, we forward all comments, $T$, to the trained network, and extract all the predictions, $Y = \{y_c | c \in T\}$. Afterwards, we initialize a vector $\overline{a}$ of size $n$, where $\overline{a}_i = |\{y | y \in Y \land y = i\}|$, for $i = (1, \dots, n)$. This vector represents the number of times each author is predicted for all comments of $a$.

\mysubsection{Graph Attention Network.}
In addition to looking at the annotator's embedding individually, we also try to model interactions between them. Let $ A = \{ a_1, \dots a_n\}$  be the set of $n$ annotators. We construct a graph  $ \mathcal{G} = (V, E)$ where the set of nodes $V$ is the set of annotators, and $E = \{e_1 \dots e_k\}$ is the set of edges, such that $e_i = (v_x, v_y)$ is an undirected edge between two nodes, and an annotator $a_x$ and $a_y$ have commented on the same situation. The final constructed graph contains 13K nodes and 1.3M edges. Graph neural networks have made significant improvements across various tasks, such as hate speech detection, misinformation spreading, suicide detection, and question answering \cite{mishra-etal-2019-snap,Chandra-2020Gra,kacupaj-etal-2021-conversational,sawhney-etal-2021-suicide,lrec_factoid,plepi-etal-2022-temporal}. There are several architectures, however, annotators have different numbers of interactions between each other. Hence, to model the influence of between nodes, we use graph attention networks (GAT; \citet{velickovic2018graph}). GAT has an attention mechanism which attends to the neighborhood of each node, and gives an importance score to the connections. 

\begin{table*}[]
    \centering
    \small
    \begin{tabular}{rcccccc}
        \toprule
         & \multicolumn{2}{c}{\textbf{No Disjoint}} & \multicolumn{2}{c}{\textbf{Situations}} & \multicolumn{2}{c}{\textbf{Authors}} \\
        \textbf{Annotator Model} & \textbf{Accuracy} & \textbf{F1} & \textbf{Accuracy} & \textbf{F1} & \textbf{Accuracy} & \textbf{F1} \\
        % \midrule
        % Majority & 70.4 & 50.0 & 71.4 & 50.0 & 70.2 & 50.0 \\
        \midrule
        Averaging Embeddings & \textbf{86.1} & \textbf{83.3} & 66.5 & 56.2 & \textbf{86.0} & \textbf{83.2} \\
        Priming & 83.9 & 80.6 & \textbf{69.6} & 52.9 & 70.2 & 41.2 \\
        Authorship Attribution & 85.5 & 82.4 & 68.4 & \textbf{56.1} & 85.2 & 82.3 \\
        Graph Attention Network & 86.0 & 83.0 & 67.8 & 54.4 & 85.6 & 82.7 \\
        Author ID & 85.1 & 82.1 & 66.7 & 55.0 & 84.3 & 81.1\\

        \bottomrule
    \end{tabular}
    \caption{Accuracy and macro F1 scores as percentages for each split method. Bolded numbers are the best results for each column and significantly outperform the next best model ($p<0.0002$ for situations, $p<0.002$ for authors, and $p<0.0004$ for the no disjoint set, with paired permutation test).}%Situations and authors are disjoint across splits for the latter two respective column pairs, whereas the first is split by neither, meaning some authors and situations (but not verdicts) overlap across splits.
    \label{tab:main_results}
\end{table*}

% \begin{table}[]
%     \centering
%     \small
%     \begin{tabular}{rcccccc}
%         \toprule
%         \textbf{Interpretation Model} & \textbf{Accuracy} & \textbf{F1} \\
%         \midrule
%         Text Only (SBERT) & 83.0 & 79.0 \\
%         \citet{botzer-etal-2022-analysis} & 78.4 & 71.6 \\
%         \midrule
%         Averaging Embeddings & 83.4 & 79.5 \\
%         Priming & 80.6 & 75.5 \\
%         Authorship Attribution & \textbf{84.2} & \textbf{79.7} \\
%         Graph Attention Network & 83.3 & 78.7 \\
%         Author ID & 84.1 & 79.5 \\
%         \bottomrule
%     \end{tabular}
%     \caption{Accuracy and macro F1 scores as percentages for the situation split. The best model is bolded and significantly better than the SBERT text only baseline ($p<0.0003$, paired permutation test)}
%     \label{tab:with_verdicts}
% \end{table}

\section{Experiments}
We experiment with four personalization methods for annotator modeling and two situation text baselines for a secondary task of personalized verdict interpretation.

\mysubsection{SBERT (text only interpretation model)}: As our base model, we finetune SBERT on the binary task of predicting the verdict, given the comment and the situation title.

\mysubsection{JudgeBERT (text only interpretation model)}:
We compare our personalized models to JudgeBert~\cite{botzer-etal-2022-analysis}, a recent model that was developed to study moral judgements, and reported the highest performance of the models discussed in \S\ref{sec:related_work}. Though our novel task setup does not have an existing baseline to directly compare to, this comparison, which does use the verdict text, serves as a point of reference.

\mysubsection{Averaging Embeddings}: We finetune the SBERT base model, and add an additional layer to concatenate the text representations with annotators representations, using the initial annotator representations computed from SBERT for Annotators \S\ref{subsec:avg}.

\mysubsection{Priming}: This model is the same as the base model, but the input text is different. The SBERT base model is finetuned on the binary task of predicting the verdict, given the situation title, the sampled text from each annotator, and the comment in the interpretation model case. 

\mysubsection{Author Attribution}: In this setup we have the same architecture as averaging embeddings, however, the initial annotator representations are generated using the author attribution model.

\mysubsection{Graph Attention Network}: In addition to the SBERT fine-tuning over the comment and the situation title, we train a GAT model to learn the annotator representations. 

\mysubsection{Author ID}: This model appends only an author-specific ID to each input. This approach is similar to the common ID-only personalization and annotator modeling approaches discussed in \S\ref{sec:related_work}.
%This model is similar to priming, however, instead of appending text which is sampled from an annotator's past posts, we only append his ID. %Note add reference. 

We train our models for 10 epochs, with the Adam optimizer, using initial learning rate $1e-4$, and focal loss \cite{8417976} to cope with class imbalance. As our base SBERT model, we use DistilRoBERTa~\cite{Sanh2019DistilBERTAD}, with dimension 768 and maximum length of 512. For the priming method, we sample $m = 100$. 
Moreover, we set $d = 768$ in the author attribution model, and train three different networks depending on the number of authors for the corresponding training split. The model is trained for 100 epochs, with the Adam optimizer, using the initial learning rate $1e-5$. 
Our experiments are run on a single NVIDIA A100 40GB GPU with an average running time (training + inference) of around one hour. 

\subsection{Three Splits}
We split the data in three ways. The first is randomly splitting verdicts into train, validation, and test. This involves two confounds; the same situations and the same authors can occur in multiple splits. Our dataset contains authors who comment on many situations, providing a verdict. A graph containing nodes corresponding to authors and posts and edges representing annotators who comment on a post, is fully connected. It is therefore not possible to remove both confounds at once without removing edges, reducing the data size, which introduces a new confound. Instead, we examine two additional splits, each controlling for one of the two confounds. The situation and author splits have disjoint sets of situations and authors respectively, across train, validation and test.

\begin{table*}[]
    \centering
    \small
    \begin{tabular}{rcccccc}
        \toprule
         & \multicolumn{2}{c}{\textbf{No Disjoint}} & \multicolumn{2}{c}{\textbf{Situations}} & \multicolumn{2}{c}{\textbf{Authors}} \\
        \textbf{Interpretation Model} & \textbf{Accuracy} & \textbf{F1} & \textbf{Accuracy} & \textbf{F1} & \textbf{Accuracy} & \textbf{F1} \\
        \midrule
        Text Only (SBERT) & 91.7 & 90.0 & 83.0 & 79.0 & 91.2 & 89.5 \\
        \citet{botzer-etal-2022-analysis} & 89.2 & 87.0 & 78.4 & 71.6 & 84.6 & 82.2 \\
        \midrule
        Averaging Embeddings & 91.7 & 90.0 & 83.4 & 79.5 & \textbf{91.7} & \textbf{90.0} \\
        Priming & 90.9 & 89.0 & 80.6 & 75.7 & 89.7 & 87.4 \\
        Authorship Attribution & \textbf{91.9} & \textbf{90.1} & \textbf{84.2} & \textbf{79.7} & 68.8 & 64.3 \\
        Graph Attention Network & 91.5 & 89.8 & 83.3 & 78.7 & 91.4 & 89.7 \\
        Author ID & 91.5 & 89.7 & 84.1 & 79.5 & 91.2 & 89.4 \\

        \bottomrule
    \end{tabular}
    \caption{Accuracy and macro F1 scores as percentages for each split method. Situations and authors are disjoint across splits for the latter two respective column pairs, whereas the first is split by neither, meaning some authors and situations (but not verdicts) overlap across splits. The best models are bolded and in the situation and author splits are significantly better than the SBERT baseline ($p<0.0003$, paired permutation test), though the result without disjoint splits was not significant.}
    \label{tab:with_verdicts}
\end{table*}

\subsection{Results}

Our main findings are in Table \ref{tab:main_results}. We find that the performance of models is similar when there are no disjoint sets across splits as when splitting by authors, with the exception of priming, which greatly suffers from not having the same authors to train on. Generalizing to new situations proved the most difficult, suggesting that having experience with interpreting specific situations is more helpful than having experience interpreting specific authors.

When comparing to a majority baseline, the accuracy is 70\% for both the author and no-disjoint splits, and 71\% for situations. The macro F1 baseline is 50\%. In all cases, we outperform the majority baseline, except for situation split accuracy. Although accuracy on the situation split is low overall, the macro F1 is still higher than the baseline. In preliminary experiments, we also tried training models with the full situation text (i.e. the full Reddit post), and found accuracy was slightly higher but F1 was lower.

We are able to compare personalization methods for this challenging task and find that priming has the highest accuracy in the situation split, while averaging embeddings and the authorship attribution approach consistently high accuracy and F1 scores. The low performance of priming is similar in the author split, which is close to the baseline, suggesting that priming often does not provide a useful signal to the model. In addition, we notice that using only the author ID as an additional token in the text, is still better than priming, which shows that using randomly sampled text from the authors might sometimes be misleading. Averaging embeddings proved to be very effective considering the simplicity of the method compared to authorship attribution and the graph attention network. In contrast to \citet{welch-etal-2022-leveraging}, we found that authorship attribution representations can scale to a large number of users by learning a projection layer to reduce it to a similar size as the text encoding. Contrary to \citet{king-cook-2020-evaluating}, who found that priming outperformed other methods in relatively low data settings (like ours), we find that it underperforms other methods at the verdict-level.
Moreover, adding a GAT offers lower improvement than averaged annotator embeddings. This is contrary to previous work~\cite{plepi-flek-2021-perceived-intended-sarcasm}, where adding the GAT layer yielded improvements. This may be due to different social media data and interactions (they used Twitter data).

Our results using the verdict text are shown in Table \ref{tab:with_verdicts}. Although this is a separate task, with a goal of interpreting an author's verdict rather than predicting it, it does provide additional insight. We find that our models greatly outperform previous work and future work should consider SBERT as a baseline without personalized features. We also find that the personalized methods outperform the text only baseline except for the priming method.  Authorship attribution often performs best, though averaging embeddings outperforms other methods on the authors split.

\begin{table}[]
    \centering
    \small
    \begin{tabular}{rccc}
        \toprule
        \textbf{Annotator Model} & \textbf{Distant} & \textbf{Close} & \textbf{Family} \\
        \midrule
        Averaging Embeddings & 59.2 & 62.1 & 65.7 \\
        Priming & 61.6 & 64.6 & 67.9 \\
        Authorship Attribution & 59.7 & 62.1 & 64.5 \\
        Graph Attention Network & 58.6 & 61.2 & 66.3 \\
        Author ID & 58.9 & 61.2 & 64.5 \\
        \bottomrule
    \end{tabular}
    \caption{Macro F1 scores for performance on the situation split. The F1 scores are first computed by the situations and then averaged across those. The results show that when the relationship between people in conflict is distant (e.g. co-workers, strangers), personalization does not help (50\% baseline), but the closer the relationship (e.g. friends, family), the more personalization helps.}
    \label{tab:cluster_breakdown}
\end{table}

\section{Analysis \& Discussion}

Here we attempt to understand how our models perform with respect to individuals, demographic groups, available data, and the type of task.

\subsection{Performance Across Tasks} %clusters=tasks

We further analyzed the performance of our methods with respect to the clusters from \S\ref{sec:clustering}. We use the situation split with no verdicts, as this most clearly demonstrates performance.
% , though the results including verdicts are in Appendix~\ref{sec:extra_results}.
The macro F1 scores\footnote{The F1 scores are first computed by the situations and then averaged across those.} are shown in Table~\ref{tab:cluster_breakdown}. We see a direct correlation between the closeness of the relationship between parties in conflict and the effectiveness of personalization. This means that for relationships such as between family members or friends, personalization methods can better learn how people will judge actions in these situations. However, when relations are more distant, such as those between co-workers or strangers, personalization methods are not as capable in helping to predict judgements. This is a key insight that raises questions for future work on judgements of social norms, but more generally suggests that the effectiveness of personalized models should be considered in terms of the properties of the classification task.

\begin{table*}[]
    \centering
    \small
    \begin{tabular}{rccccccc}
        \toprule
         & \multicolumn{3}{c}{\textbf{Gender}} & \multicolumn{3}{c}{\textbf{Age}} \\
        \textbf{Annotator Model} & \textbf{Male} & \textbf{Female} & \textbf{Unknown} & \textbf{Younger} & \textbf{Older} & \textbf{Unknown} & \textbf{All} \\
        \midrule
        Averaging Embeddings & 65.8 & 65.2 & 66.7 & 65.4 & 65.9 & 66.3 & 66.3 \\
        Priming & 70.4 & 69.7 & 70.1 & 68.6 & 70.4 & 70.2 & 69.0 \\
        Authorship Attribution & 68.3 & 67.1 & 68.7 & 68.0 & 68.1 & 68.2 & 68.2\\
        Graph Attention Network & 67.5 & 67.6 & 68.1 & 67.1 & 67.5 & 68.0 & 67.0 \\
        Author ID & 66.0 & 66.3 & 66.0 & 66.2 & 66.6 & 66.0 & 67.1 \\

        \bottomrule
    \end{tabular}
    \caption{Breakdown of annotator-level accuracies for each personalization method in the situation split. We show performance independently for age and gender, using three values for each.}
    \label{tab:demo_results}
\end{table*}

\subsection{Who Personalization Helps}
% Breakdown by demographics and clusters
% box plots of performance distribution across users -- how much variance? would be interesting to see
% what is the correlation between the amount of text from a user and how well we predict their verdicts? (something like this for the situation split could be interesting)

\pgfplotsset{width=7.5cm,height=5cm,compat=1.18}
\usepgfplotslibrary{statistics}

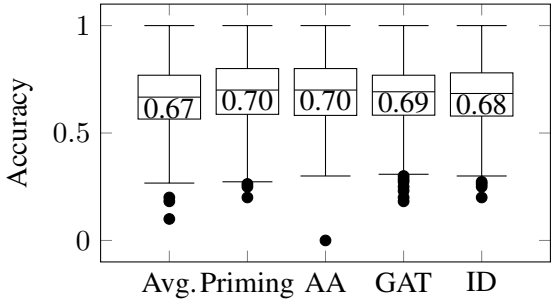
\begin{figure}[]
\begin{tikzpicture}
\begin{axis}[boxplot/draw direction=y,
            xtick={1,2,3,4,5},
            xticklabels={Avg., Priming, AA, GAT, ID},
            ylabel={Accuracy},
            ]

    \addplot [boxplot prepared={
    lower whisker=0.267, lower quartile=0.565,
    median=0.667, upper quartile=0.769,
    upper whisker=1.0},
    ] coordinates {(1,0.182) (1,0.1) (1,0.2)};
    
    \addplot [boxplot prepared={
    lower whisker=0.273, lower quartile=0.587,
    median=0.7, upper quartile=0.8,
    upper whisker=1.0},
    ] coordinates {(3,0.2) (3,0.263) (3,0.25)};
    
    \addplot [boxplot prepared={
    lower whisker=0.3, lower quartile=0.582,
    median=0.7, upper quartile=0.8,
    upper whisker=1.0},
    ] coordinates {(0,0)};
    
    \addplot [boxplot prepared={
    lower whisker=0.308, lower quartile=0.583,
    median=0.692, upper quartile=0.769,
    upper whisker=1.0},
    ] coordinates {(2,0.3) (2,0.25) (2,0.273) (2,0.286) (2,0.182) (2,0.2) (2,0.231)};
    
    \addplot [boxplot prepared={
    lower whisker=0.3, lower quartile=0.579,
    median=0.684, upper quartile=0.78,
    upper whisker=1.0},
    ] coordinates {(4,0.2) (4,0.263) (4,0.25) (4,0.273)};
    
    \node at (axis cs: 1,.617) {0.67};
    \node at (axis cs: 2,.65) {0.70};
    \node at (axis cs: 3,.65) {0.70};
    \node at (axis cs: 4,.642) {0.69};
    \node at (axis cs: 5,.634) {0.68};
\end{axis}
\end{tikzpicture}
\caption{Distribution of accuracy for the no-verdict situation split across users with at least 10 verdicts in the test set ($N=715$). Median values are in the boxes. Abbreviations: AA=Authorship Attribution, Avg.=Averaging Embeddings, GAT=Graph Attention Network, ID=Author ID.}
\label{fig:boxes_user_dist_acc}
\end{figure}
% \vskip -0.2in

To be able to better understand the impact across individuals, we plotted the distribution of accuracies for the situation split with no verdicts in Figure~\ref{fig:boxes_user_dist_acc}. We see that for our models, the variance is similar. We find that performance is much higher for some annotators than others.

To further examine differences we looked at the annotator-level performance across demographics extracted as in \S\ref{sec:extract_demos}. We use three values for each, including a value for \textit{unknown}.  The results in Table~\ref{tab:demo_results} show the accuracies for each demographic do not vary much from the overall scores for each model. Interestingly, we find that those with unknown demographics tend to have slightly higher performance. This group of individuals who are less likely to share demographic information may have something in common that is beneficial for modeling judgements, though this remains to be explored. Overall, our models show relatively fair performance across demographics, not strongly performing for one group over another, even with an uneven distribution of genders. Accordingly, when the only identifiers used for the annotators, are their IDs, the performance across demographics is at $66\%$.

\subsection{Annotator Data Volume}

When looking at annotator-level accuracy, one may wonder if it helps to have more data for an annotator during training. We tested this with the situation split using Pearson's correlation coefficient. We calculated annotator-level accuracy for each annotator and grouped them by the amount of available data points during training (up to 463). Then, we calculated the correlation between the amount of data and the mean accuracy across annotators that had that amount of training data.

Across methods on the situation split we find $r=0.19-0.22$ ($p<0.05$) with slight variance across methods. When we looked more specifically at the three tasks, finding that for the Distant task, correlation is weaker, $r=0.14-0.19$ ($p<0.08$), and we find no correlation in the Close task, $r=-0.02-0.11$ ($p<0.6$). With the Family task we find the strongest correlation with $r=0.18-0.25$ ($p<0.02$). Having more data per annotator helps with Family, but does not help as much with the more distant clusters. Interestingly, having more data seems to help more with predicting verdicts of the Distant task than it does the Close task.

\section{Conclusion}

We experimented with a recently constructed dataset with unique properties that allowed us to explore annotator modeling using personalization methods. We filtered and augmented the data to obtain a set suitable for our experiments which contains self-identified demographics. Our approach unifies calls for data perspectivism with approaches developed for large volumes of user information. Within this setting we are also able to provide a comparison of recent personalization methods.

Overall, we found that averaging embeddings provided a strong and relatively simple approach, though each model has its strengths, the authorship attribution and graph attention networks were consistently high performing across splits, while for the situation split, annotator-level accuracy was highest with the priming approach. These methods outperform the common approach of representing authors with a single ID.
As a secondary result, we found that personalized methods significantly outperformed previous work and text only baselines on the task of interpreting verdicts.

We then showed an analysis of performance across tasks, showing that the closer the relationship between parties in conflict, the more personalization helps model the judgements annotators will give. We further showed performance across demographics, showing that our methods appear unbiased in this regard, despite being trained with more data from females than males. We revealed a correlation between the amount of data from an annotator during training, and it's impact on personalization, showing that more data generally, but not always, helps.

We hope that our formalization of this task provides a path for future work in this direction and our insights for the task of predicting judgements of social norms provide meaningful first steps.
Our code is publicly available at \href{https://github.com/caisa-lab/perspectivism-personalization.git}{https://github.com/caisa-lab/perspectivism-personalization.git}

\section*{Limitations}
When extracting demographics from the history of the authors in our dataset we used all comments from January 2014 to June 2021 provided in the PushShift API. We follow the recommendations of \cite{larson-2017-gender}, and do not refer to biological sex but rather to a users self stated gender. 
After applying the extraction method, users are assigned \textit{female}, \textit{male}, or \textit{unknown}.
We acknowledge that through our approach a range of non-binary, transgender, and gender fluid people are excluded and that some users could be assigned incorrectly as our extraction is not perfect.

We faced the fact that it is not uncommon to have multiple ages or genders for a single user.
To limit the number of incorrectly captured ages, for a set of extracted ages of an author, we checked if the range is at most the range of years we extracted from and take the maximum, representing the latest age extracted.
If the set's range exceeds seven and a half years (range of time data was collected), we concluded that ages varying the most from the median of the set are outliers that originate from cases such as an author writing fictitiously.
In this case, we recursively remove the outliers, to check if the range possible but contains at lest three matches.
When multiple genders occur, we checked if one gender makes up at least 80\% of the set to be more certain in applying the label.

Our experiments pertain to data collected from Reddit, meaning that others should be aware that our findings may not generalize to corpora that are significantly different, or communities that are made up of different types of users.

%Clustering a large set of descriptions of conflict situations is a difficult task, as was shown by our manual annotation. Although our agreement was reasonable, our findings across the described relationship groups are less clear for different possible categorizations, or how people may subjectively draw the line between different groups.

% EMNLP 2022 requires all submissions to have a section titled ``Limitations'', for discussing the limitations of the paper as a complement to the discussion of strengths in the main text. This section should occur after the conclusion, but before the references. It will not count towards the page limit.  

% The discussion of limitations is mandatory. Papers without a limitation section will be desk-rejected without review.
% ARR-reviewed papers that did not include ``Limitations'' section in their prior submission, should submit a PDF with such a section together with their EMNLP 2022 submission.

% While we are open to different types of limitations, just mentioning that a set of results have been shown for English only probably does not reflect what we expect. 
% Mentioning that the method works mostly for languages with limited morphology, like English, is a much better alternative.
% In addition, limitations such as low scalability to long text, the requirement of large GPU resources, or other things that inspire crucial further investigation are welcome.

\section*{Ethics Statement}
% Scientific work published at EMNLP 2022 must comply with the \href{https://www.aclweb.org/portal/content/acl-code-ethics}{ACL Ethics Policy}. We encourage all authors to include an explicit ethics statement on the broader impact of the work, or other ethical considerations after the conclusion but before the references. The ethics statement will not count toward the page limit (8 pages for long, 4 pages for short papers).
Personalized technologies seek to model individuals, which may also be used to identify and surveil them. This relates to tasks such as author profiling~\cite{rangel2013overview}. The methods in this work could be used to build a profile of peoples views of social norms. We advocate against such uses, as they may result in discrimination or threats to intellectual freedom~\cite{richards1934dangers}.

Personalization, as presented in our paper, can be used to automatically infer people's opinions about social norms and other subjective stance, sentiment, or perception. This could be desired in some applications but could be undesired or even harmful in others.  Bias in models can cause misrepresentation and negatively impact populations \cite{blodgett-etal-2020-language, sap2021annotators}. These populations may be represented with the demographics we identify, or may result from sample bias. %Reddit's users reflect a specific population that tends to be young, male, from the USA, with internet access. 
Those not well represented in our data may be negatively affected by the application of models such as ours depending on their use. Although the implications depend on the application, we generally suggest that if such a method is used in practice, that end-users are made aware of how their data is being used and given the choice to not be a part of automated decision processes based on these inferences.

\section*{Acknowledgements}
This work has been supported by the German Federal Ministry of Education and Research (BMBF) as a part of the Junior AI Scientists program under the reference 01-S20060, the Alexander von Humboldt Foundation, and by Hessian.AI. Any opinions, findings, conclusions, or recommendations in this material are those of the authors and do not necessarily reflect the views of the BMBF, Alexander von Humboldt Foundation, or Hessian.AI. Credit for the Reddit icon in Figure~\ref{fig:reddit_example} goes to Freepik at \url{flaticon.com}.

\bibliography{anthology,rebib}
\bibliographystyle{acl_natbib}

\end{document}